\documentclass[conference]{IEEEtran}
\IEEEoverridecommandlockouts
\usepackage[table]{xcolor}
\usepackage{mathrsfs,amssymb,amsmath}
\usepackage{graphicx}
\usepackage{multirow} 
\usepackage{xspace}
\usepackage{booktabs}
\usepackage{enumitem} 
\usepackage{longtable}
\usepackage[htt]{hyphenat} 
\usepackage{dirtree}
\usepackage{eurosym}
\usepackage{dirtree}
\definecolor{vlgray}{gray}{0.92}

\definecolor{grey}{rgb}{0.75,0.75,0.75}

\usepackage{cite}
\usepackage{amsmath,amssymb,amsfonts}
\usepackage{algorithmic}
\usepackage{graphicx}
\usepackage{textcomp}
\usepackage{tikz}
\usepackage{xcolor}
\usepackage{url}
\usepackage{listings}
\usetikzlibrary{shapes}
\usepackage{xcolor}
\def\BibTeX{{\rm B\kern-.05em{\sc i\kern-.025em b}\kern-.08em
    T\kern-.1667em\lower.7ex\hbox{E}\kern-.125emX}}
    
\usepackage{eurosym}
\usepackage{booktabs}

\begin{document}

\title{An Empirical Evaluation of Two \\ General Game Systems: Ludii and RBG \\
\thanks{Funded by a \euro2m ERC Consolidator Grant (\texttt{http://ludeme.eu}).}}

\author{\IEEEauthorblockN{{\'E}ric Piette, Matthew Stephenson, Dennis J.N.J. Soemers and Cameron Browne}
\IEEEauthorblockA{\textit{Department of Data Science and Knowledge Engineering} \\
\textit{Maastricht University}\\
Maastricht, the Netherlands \\
\texttt{\{eric.piette,matthew.stephenson,dennis.soemers,cameron.browne\}@maastrichtuniversity.nl}}
}

\maketitle

\begin{abstract}
Although General Game Playing (GGP) systems can facilitate useful research in Artificial Intelligence (AI) for game-playing, they are often computationally inefficient and somewhat specialised to a specific class of games. However, since the start of this year, two General Game Systems have emerged that provide efficient alternatives to the academic state of the art -- the Game Description Language (GDL). In order of publication, these are the Regular Boardgames language (RBG), and the Ludii system.
This paper offers an experimental evaluation of Ludii. Here, we focus mainly on a comparison between the two new systems in terms of two key properties for any GGP system: simplicity/clarity (e.g. human-readability), and efficiency.
\end{abstract}


\section{Introduction}
Playing games has been such a long standing aspect of human culture that it can certainly be considered an integral part of our history. Strategic games can provide a complex challenge for even the most skilled human player, often requiring a precise and forethought sequence of response actions to win.
As an alternative to creating specialised algorithms for specific games like Chess \cite{stockfish17}, or Go \cite{silver16}, developing artificial agents capable of playing a broad variety of games is the goal of \textit{General Game Playing} (GGP) \cite{pitrat68}. The modern era of GGP was launched in 2005 with the annual International General Game Playing Competition (IGGPC) announced by the Stanford's Logic Group \cite{genesereth05}. 

The General Game System GGP-BASE \footnote{GGP-BASE \url{https://github.com/ggp-org/ggp-base}} using the \textit{Game Description Language} (GDL) \cite{love08} has become the standard for academic research in GGP. 
Games in GDL are described in terms of simple instructions based on first-order logic clauses, designed for deterministic games with perfect information. Extensions to this language also allow for imperfect information and epistemic games \cite{schiffel14,thielscher17}. The general intelligence required by GDL agents has led to several important research contributions \cite{bjornsson16}, with one of the most popular and effective techniques being Monte Carlo tree search \cite{finnsson08,finnsson10}, while the last winner of the IGGPC uses an approach based on constraint programming and symmetry detection \cite{koriche17}.

However, defining games using GDL currently has several problems. 
Writing and debugging game descriptions requires an adept understanding of first order logic, making such tasks difficult for people who are not computer scientists or mathematicians.
The equipment and rules for each game are also typically interconnected, and the structural aspects of each game, such as the board, deck, or arithmetic operators, must be explicitly defined from scratch each time. This makes game creation a lengthy and time-consuming process.
Even small modifications to existing games, such as changing the size of the board, require many lines of code to be changed or added.
Processing such descriptions is also computationally expensive as it requires logic resolution, making the language difficult to integrate with other external applications and limiting the potential of GDL outside of game AI.

Due to these limitations of GDL, an alternative GGP language is useful which allows games to be defined in an easier and human understandable manner, whilst also being efficient enough to facilitate future AI research in many other areas of GGP, like developing universal agents, learning, procedural content generation, or game analysis. Apart from AI research, developing such software can provide an efficient tool for work in related fields such as game design, history, or education.




\section{Two Alternative General Game Systems}

Recently, two alternatives to GDL have been proposed: the Regular Boardgames language (RBG) \cite{kowalski19}, and Ludii \cite{Piette19}.

\subsection{Regular Boardgames (RBG)} 

The idea behind RBG comes from an initial work proposed by \cite{Bjornsson12} for using a regular language to encode the movement of pieces for a small subset of chess-like games called Simple Board Games. However, as the allowed expressions are simplistic and applied only to one piece at a time, it cannot express any non-standard behavior. RBG \cite{kowalski19} extended and updated this idea to be able to describe the full range of deterministic board games. 

\begin{figure}[!t]
\footnotesize
\lstset{numbers=left, numberstyle=\tiny, stepnumber=1,%
numberfirstline=false, numbersep=5pt,basicstyle=\ttfamily}
\begin{lstlisting}
<@\textcolor{red}{\#players}@> = white(100), black(100)
<@\textcolor{red}{\#pieces}@> = e, w, b, x
<@\textcolor{red}{\#variables}@> =
<@\textcolor{red}{\#board}@> = rectangle(up,down,left,right,
         [e, e, e, b, e, e, b, e, e, e]
         [e, e, e, e, e, e, e, e, e, e]
         [e, e, e, e, e, e, e, e, e, e]
         [b, e, e, e, e, e, e, e, e, b]
         [e, e, e, e, e, e, e, e, e, e]
         [e, e, e, e, e, e, e, e, e, e]
         [w, e, e, e, e, e, e, e, e, w]
         [e, e, e, e, e, e, e, e, e, e]
         [e, e, e, e, e, e, e, e, e, e]
         [e, e, e, w, e, e, w, e, e, e])

<@\textcolor{red}{\#anySquare}@> = ((up* + down*)(left* + right*))

<@\textcolor{red}{\#queenShift}@> = (
      (up left {e}) (up left {e})* +
      (up {e}) (up {e})* +
      (up right {e}) (up right {e})* +
      (left {e}) (left {e})* +
      (right {e}) (right {e})* +
      (down left {e}) (down left {e})* +
      (down {e}) (down {e})* +
      (down right {e}) (down right {e})*
    )

<@\textcolor{red}{\#turn}@>(piece; player) = (
    anySquare {piece} [e]
    queenShift
    [piece]
    queenShift
    ->> [x]
  )

<@\textcolor{red}{\#rules}@> = ->white (
    turn(w; white) 
        [$ white=100, black=0] -> black
    turn(b; black) 
        [$ white=0, black=100] -> white
  )*
\end{lstlisting}
\caption{Game description of Amazons with RBG.}
 \label{fig:breakthroughRBG}
\end{figure}

The RBG system uses a low-level language given as an input for programs (agents and game manager), which is easy to process, and a high-level language, which allows for more concise and human readable descriptions. The high level version can be converted to the low level version in order to provide the two main aspects of a GGP system: human-readability and efficiency for AI programs. The technical syntax specification of RBG can be found in \cite{Kowalski18}. Thanks to this distinction between two languages, it is possible to model complex games (e.g. Amazons, Arimaa or Go), and apply the more common AI methods (minimax, Monte-Carlo search, Reinforcement Learning, etc.) to them. Indeed, in the previous languages used for GGP (including the academic state of the art GDL), it was difficult to model any complex games and impossible to play or reason on any of them in a reasonable amount of time. As an example, Figure \ref{fig:breakthroughRBG} presents a complete description of Amazons for the RBG system. RBG has been demonstrated to be universal for the class of finite deterministic games with full information, and more efficient than the state of the art using GDL \cite{kowalski19}. 

\subsection{Ludii}

Within the context of the Digital Ludeme Project\footnote{Digital Ludeme Project: \url{www.ludeme.eu}} \cite{browne18} -- which aims to model the world's traditional strategy games in a single, playable digital database -- using GDL as the game description language to model games was not feasible due to the inherent complexity of modelling games with first-order logic, and the difficulty of integrating this language with other external applications. For this reason, a new General Game system based on Browne's thesis \cite{browne09} and the notion of \textit{ludemes}, called Ludii \cite{Piette19} was implemented. 

Ludemes are the conceptual elements of a game. In Ludii, games are composed of ludemes in order to distinguish the game's form (rules and equipment) and its function (its emergent behaviour). An important benefit of the ludemic approach is that it encapsulates key game concepts and gives them meaningful labels. Thanks to this property, it is possible to use the class grammar approach \cite{BrowneB16} to derive the game description language directly from the class hierarchy of the underlying source code library used by the system. For this reason, the game description language used by Ludii is automatically generated from the constructors in the class hierarchy of the Ludii source code. Game descriptions expressed in the grammar are automatically instantiated back into the corresponding library code for compilation, giving a guaranteed 1:1 mapping between the source code and the grammar.

\begin{figure}[!t]
\footnotesize
\lstset{numbers=left, numberstyle=\tiny, stepnumber=1,%
numberfirstline=false, numbersep=5pt,basicstyle=\ttfamily}
\begin{lstlisting}
(<@\textcolor{red}{game}@> "Amazons"  
 (<@\textcolor{red}{mode}@> 2)  
 (<@\textcolor{red}{equipment}@> 
  { 
  (chessBoard 10) 
   (queen 
     Each 
     (slide (in (to) (empty)) 
      (then (replay))
     )
   )
   (dot None)
  }
 )
 
 (<@\textcolor{red}{rules}@> 
  (<@\textcolor{red}{start}@>
   { 
    (place "Queen1" {3 6 30 39})
    (place "Queen2" {60 69 93 96})
   }
  )
    
  (<@\textcolor{red}{play}@> 
   (if (even (turn))
  	(byPiece)
   	(shoot (in (to) (empty)) "Dot0"))
  )
    
  (<@\textcolor{red}{end}@> 
   (stalemated (mover)) 
   (result (next) Win) 
  )  
 )
)
\end{lstlisting}
\caption{Game description of Amazons with Ludii.}
 \label{fig:breakthroughLUDII}
\end{figure}

This approach produces an efficient tool to model, play and analyse any strategical games as a structure of ludemes. Moreover, it is also possible to associate any kind of information like the kind of game state (e.g. stacking, boardless, stochastic, etc) to each ludeme in order to optimise the reasoning on games and to provide a playable interface for each game. Ludii makes its programming language the game description language. It can theoretically support any rule, equipment or behaviour that can be programmed, but the implementation details are hidden from the user, who only sees the simplified grammar which summarises the code to be called. A more complete description of the Ludii system is given in \cite{Piette19}, and an example, Figure \ref{fig:breakthroughLUDII} presents a complete description of Amazons for Ludii. 

Like RBG, Ludii has been demonstrated to be universal for the class of finite deterministic games with full information, and more efficient than GDL \cite{Piette19}. However, due to the novelty of the two new GGP systems, they were not yet compared to each other. We remedy this in the next section.

\section{Experiments}

The two new GGP Systems -- as most other GGP systems -- use MCTS as the core method for AI move planning, which has proven to be a superior approach for general games in the absence of domain specific knowledge \cite{finnsson10}. 
MCTS playouts require fast reasoning engines to achieve the desired number of simulations. 
Hence, we use {\it flat Monte Carlo} playouts as the metric for comparing the efficiency between Ludii and RBG. 

In order to make a comparison between the human-readability of the two systems we use two main criteria: clarity and simplicity. Clarity refers to the degree to which game descriptions would be self-explanatory to non-specialist readers, and simplicity refers to the ease with which game descriptions can be created and modified, and can be estimated by the number of tokens required to define games.


\subsection{Setup}

In the following comparison, we compare RBG and Ludii based on the numbers of tokens used to describe each game, and the number of random playouts obtained per second by each of them. All experiments were conducted on a single core of an Intel(R) Core(TM) i7-8650U CPU @ 1.90 GHz, 2112 MHz with 16GB RAM, spending 10 min per test.
The RBG system is implemented in C++ and compiled with g++ 8.3 and Ludii is implemented in Java and compiled with the Java SE Development Kit 11. The RBG system provides an interpreter and a compiler to perform reasoning, consequently we compare Ludii to each of them.

\subsection{Results}

The results of our experiments for the same set of games implemented on the two systems are shown in Table \ref{table:results}.
The left section of the table is dedicated to the number of tokens used to describe the games, and the right section to the number of playouts obtained per second (the highest number of playouts obtained for each game is coloured in blue). 
The rightmost columns of each section shows the rate of Ludii on RBG. Note that results are also given for GDL in order to highlight the gap between it and the new alternatives.

{
\centering
\begin{table*}[!t]
\caption{An experimental comparison between GDL, RBG, and Ludii (rate = Ludii / RBG).}
\label{table:results}
   \begin{center}
\begin{tabular}{@{}l|ccccc|ccccccc@{}}
\toprule
\bf  & \bf {\tt \bf  } & \bf {\tt \bf } &\bf {\tt \bf } & \bf {\bf } & \bf {\bf } & \bf {\bf }  & \bf {\tt \bf } & \bf {\tt \bf RBG} & \bf {\tt \bf RBG} &\bf {\tt \bf } & \bf {\bf Rate} & \bf {\bf Rate} \\
\bf Game & \bf {\tt \bf  GDL} & \bf {\tt \bf RBG} &\bf {\tt \bf Ludii} & \bf {\bf Rate} & \bf {\bf } & \bf {\bf }  & \bf {\tt \bf GDL} & \bf {\tt \bf Interpreter} & \bf {\tt \bf Compiler} &\bf {\tt \bf Ludii} & \bf {\bf Interpreter} & \bf {\bf Compiler}\\
\midrule
\rowcolor{vlgray} \multicolumn{1}{c}{} & \multicolumn{4}{c}{{\bf Number of tokens}} & \multicolumn{4}{c}{} & \multicolumn{2}{c}{{\bf Playouts per second}} & \multicolumn{2}{c}{}\\
Amazons & 1158 & 195 & 51 & 0.26 & & & 185 & 307 & 625 & \textcolor{blue}{4,349} & 14.17 & 6.96\\
Arimaa  & $\times$ & 735 & 359 & 0.49 & & & $\times$ & 0.01 & 0.11 & \textcolor{blue}{714} & 71,400 & 6,490.9 \\
Breakthrough & 670 & 134 & 65 & 0.49 & & & 1,123 & 4,962 & \textcolor{blue}{16,694} & 4,741 & \textcolor{red}{0.96} & \textcolor{red}{0.28} \\
Chess & 4392 & 641 & 186 & 0.29 & & & 0.06 & 79 & 714 & \textcolor{blue}{720} & 9.11 & 1.01 \\
Chinese Checkers & $\times$ & 418 & 243 & 0.58 & & & $\times$ & 232 & 1090 & \textcolor{blue}{1,105} & 4.76 & 1.01\\
Connect-4 & 751 & 155 & 31 & 0.20 & & & 13,664 & 41,897 & 84,124 & \textcolor{blue}{94,077} & 2.25 & 1.12\\
English Checkers & 1282 & 263 & 161 & 0.61 & & & 872 & 2,963 & \textcolor{blue}{14,286} & 8,135 & 2.75 & \textcolor{red}{0.57} \\
Double Chess & $\times$ & 790 & 202 & 0.26 & & & $\times$ & 6 & 50 & \textcolor{blue}{81} & 13.5 & 1.62 \\
Gomoku & 514 & 324 & 21 & 0.06 & & & 927 & 1,330 & 2212 & \textcolor{blue}{42,985} & 32.32 & 19.43 \\
Hex & $\times$ & 245 & 81 & 0.33 & & & $\times$ & 2741 & 5787 & \textcolor{blue}{11,077} & 4.04 & 1.91\\
International Checkers & $\times$ & 498 & 244 & 0.49 & & & $\times$ & 262 & 1,941 & \textcolor{blue}{3,444} & 13.15 & 1.77\\
Reversi & 894 & 311 & 103 & 0.33 & & & 203 & 1468 & 2,012 & \textcolor{blue}{2,081} & 1.42 & 1.03\\
The Mill Game & $\times$ & 296 & 103 & 0.15 & & & $\times$ & 1,063 & 7,423 & \textcolor{blue}{72,734} & 68.42 & 9.80\\
Tic-Tac-Toe & 381 & 101 & 25 & 0.25 & & & 85,319 & 139,312 & 400,000 & \textcolor{blue}{535,294} & 3.84 & 1.34\\
\bottomrule
\end{tabular}
\normalsize
\end{center} 
\end{table*}
}

\section{Discussion}

\subsection{Clarity and Simplicity}
We argue that describing a game with the ludemic approach is typically simpler than using regular expressions as in RBG. Indeed, the number of tokens used by Ludii to describe a game is always much smaller than RBG; the highest rate is 0.61 for English Checkers and the lowest rate is 0.06 for Gomoku. On average Ludii needs a third of the number of tokens used by RBG to model a game. 

In Ludii, the Java classes that define each ludeme use convenient definitions for the concepts involved, which provide meaningful names for each class and are aligned with human perception. For example, Container sub-classes include Board, Hand, Deck, etc., move rules include Slide, Step, etc., and game properties include stacking, hidden information, etc. These are common terms that most game players and designers would recognise and understand.
In RBG, the high-level version allows predefined functions to generate regular shaped boards, and it is possible to define a game using only meaningful names with the assistance of some basic macros.

While the clarity of a game description is subjective, we believe that most -- in particular non-expert users -- would find Ludii game descriptions to be more clear in general than RBG descriptions. See Figures \ref{fig:breakthroughRBG} and \ref{fig:breakthroughLUDII} for a comparison of the game descriptions for Amazons. In RBG, high-level concepts such as the movement of a queen can be summarised in a \texttt{\#queenShift} macro, but the macro is game-specific and its low-level details are still present in the same game description. These low-level details can be difficult to understand, especially without prior knowledge of regular expressions. Similar details are completely hidden in the Ludii game description, with the move rules being encoded in ludemes with easily-understandable names such as ``slide'' and ``shoot''. Additionally, there is a clear demarcation between starting rules, playing rules, and end conditions in Ludii.

\subsection{Efficiency}

Ludii outperforms RBG in terms of playouts per second in all evaluated games, except for Breakthrough (where RBG's interpreter and compiler are both more efficient), and English Checkers (where only RBG's compiler performs better).

For the majority of the games, the number of playouts per second computed by Ludii is significantly higher than by the RBG interpreter. If we focus on complex games, like Amazons, Chess or International Checkers, Ludii is at least $10$ times faster. Moreover for Arimaa, the RBG interpreter cannot compute a single playout per second, in contrast to $700$ playouts per second for Ludii. In the original RBG paper, a variant of Arimaa is used with a fixed position for all the pieces on the initial state in order to minimise the branching factor implied by that game, but Ludii does not need that to reason on that game. 


The RBG compiler is more efficient than the RBG interpreter in terms of reasoning and the rate between it and Ludii is lower. When using the compiler version, RBG exceeds Ludii on English Checkers as well as Breakthrough. 
For all the other games, Ludii still provides speedups over the RBG compiler.
The high performance of RBG on Breakthrough is likely due to the use of only basic onboard operations in RBG's Breakthrough description.

\section{Conclusion}

The ludemic General Game system Ludii and the Regular Boardgames system are two efficient alternatives which outperform the current standard for academic AI research into GGP in terms of reasoning and human-readability. In term of efficiency, Ludii outperforms the RBG interpreter in all evaluated games except one, and compared to the RBG compiler, it provides some better results in all evaluated games except two.

We argue that Ludii games are simpler to model, using fewer tokens to describe them than RBG. We also believe that Ludii game descriptions are easier to read, largely thanks to meaningful ludeme names, whereas RBG requires the use of game-specific macros to describe higher-level concepts.


\section*{Acknowledgment.}

This research is part of the European Research Council-funded Digital Ludeme Project (ERC Consolidator Grant \#771292) run by Cameron Browne at Maastricht University's Department of Data Science and Knowledge Engineering. 

\bibliographystyle{IEEEtran}
\bibliography{IEEEabrv,References}

\begin{thebibliography}{10}
\providecommand{\url}[1]{#1}
\csname url@samestyle\endcsname
\providecommand{\newblock}{\relax}
\providecommand{\bibinfo}[2]{#2}
\providecommand{\BIBentrySTDinterwordspacing}{\spaceskip=0pt\relax}
\providecommand{\BIBentryALTinterwordstretchfactor}{4}
\providecommand{\BIBentryALTinterwordspacing}{\spaceskip=\fontdimen2\font plus
\BIBentryALTinterwordstretchfactor\fontdimen3\font minus
  \fontdimen4\font\relax}
\providecommand{\BIBforeignlanguage}[2]{{%
\expandafter\ifx\csname l@#1\endcsname\relax
\typeout{** WARNING: IEEEtran.bst: No hyphenation pattern has been}%
\typeout{** loaded for the language `#1'. Using the pattern for}%
\typeout{** the default language instead.}%
\else
\language=\csname l@#1\endcsname
\fi
#2}}
\providecommand{\BIBdecl}{\relax}
\BIBdecl

\bibitem{stockfish17}
T.~Romstad, M.~Costalba, J.~Kiiski, and G.~Linscott, ``Stockfish: Strong open
  source chess engine,'' \url{https://stockfishchess.org/}, 2008.

\bibitem{silver16}
D.~Silver, A.~Huang, C.~J. Maddison, A.~Guez, L.~Sifre, G.~van~den Driessche,
  J.~Schrittwieser, I.~Antonoglou, V.~Panneershelvam, M.~Lanctot, S.~Dieleman,
  D.~Grewe, J.~Nham, N.~Kalchbrenner, I.~Sutskever, T.~P. Lillicrap, M.~Leach,
  K.~Kavukcuoglu, T.~Graepel, and D.~Hassabis, ``Mastering the game of go with
  deep neural networks and tree search,'' \emph{Nature}, vol. 529, no. 7587,
  pp. 484--489, 2016.

\bibitem{pitrat68}
J.~Pitrat, ``Realization of a general game-playing program,'' in \emph{IFIP
  Congress}, 1968, pp. 1570--1574.

\bibitem{genesereth05}
M.~R. Genesereth, N.~Love, and B.~Pell, ``General game playing: Overview of the
  {AAAI} competition,'' \emph{{AI} Magazine}, vol.~26, no.~2, pp. 62--72, 2005.

\bibitem{love08}
N.~Love, T.~Hinrichs, D.~Haley, E.~Schkufza, and M.~Genesereth, ``General game
  playing: Game description language specification,'' 2008.

\bibitem{schiffel14}
S.~Schiffel and M.~Thielscher, ``Representing and reasoning about the rules of
  general games with imperfect information,'' \emph{Journal of Artificial
  Intelligence Research}, vol.~49, pp. 171--206, 2014.

\bibitem{thielscher17}
M.~Thielscher, ``{GDL-III}: A description language for epistemic general game
  playing,'' in \emph{Proceedings of the Twenty-Sixth International Joint
  Conference on Artificial Intelligence}, 2017, pp. 1276--1282.

\bibitem{bjornsson16}
Y.~Bj{\"o}rnsson and S.~Schiffel, ``General game playing,'' in \emph{Handbook
  of Digital Games and Entertainment Technologies}.\hskip 1em plus 0.5em minus
  0.4em\relax Springer, 2016, pp. 1--23.

\bibitem{finnsson08}
H.~Finnsson and Y.~Bj{\"o}rnsson, ``Simulation-based approach to general game
  playing,'' in \emph{The Twenty-Third AAAI Conference on Artificial
  Intelligence}, 2008, pp. 259--264.

\bibitem{finnsson10}
------, ``Learning simulation control in general game-playing agents,'' in
  \emph{The Twenty-Fourth AAAI Conference on Artificial Intelligence}, 2010,
  pp. 954--959.

\bibitem{koriche17}
F.~Koriche, S.~Lagrue, {\'E}.~Piette, and S.~Tabary, ``Constraint-based
  symmetry detection in general game playing,'' in \emph{Proceedings of the
  Twenty-Sixth International Joint Conference on Artificial Intelligence},
  2017, pp. 280--287.

\bibitem{kowalski19}
J.~Kowalski, M.~Maksymilian, J.~Sutowicz, and M.~Szykula, ``Regular
  boardgames,'' in \emph{The Thirty-Third AAAI Conference on Artificial
  Intelligence}, 2019.

\bibitem{Piette19}
{\'{E}}.~Piette, D.~J. N.~J. Soemers, M.~Stephenson, C.~F. Sironi, M.~H.~M.
  Winands, and C.~Browne, ``Ludii - the ludemic general game system,''
  \emph{CoRR}, vol. abs/1905.05013, 2019.

\bibitem{Bjornsson12}
Y.~Bj\"{o}rnsson, ``Learning rules of simplified boardgames by observing,'' in
  \emph{Proceedings of the Twentieth European Conference on Artificial
  Intelligence}, 2012, pp. 175--180.

\bibitem{Kowalski18}
\BIBentryALTinterwordspacing
J.~Kowalski, M.~Mika, J.~Sutowicz, and M.~Szykula, ``Regular boardgames,''
  2018. [Online]. Available: \url{http://arxiv.org/abs/1706.02462}
\BIBentrySTDinterwordspacing

\bibitem{browne18}
C.~Browne, ``Modern techniques for ancient games,'' in \emph{2018 {IEEE}
  Conference on Computational Intelligence and Games}.\hskip 1em plus 0.5em
  minus 0.4em\relax IEEE, 2018, pp. 490--497.

\bibitem{browne09}
\BIBentryALTinterwordspacing
C.~B. Browne, ``Automatic generation and evaluation of recombination games,''
  Ph.D. dissertation, Queensland University of Technology, 2009. [Online].
  Available: \url{https://eprints.qut.edu.au/17025/}
\BIBentrySTDinterwordspacing

\bibitem{BrowneB16}
------, ``A class grammar for general games,'' in \emph{Advances in Computer
  Games}, ser. LNCS, vol. 10068, 2016, pp. 167--182.

\end{thebibliography}

\end{document}